\pdfoutput=1
\documentclass[11pt]{article}
\usepackage[final]{acl}
\usepackage{longtable}
\usepackage{enumitem}
\usepackage{geometry}
\usepackage{graphicx}
\usepackage{adjustbox}
\usepackage{amssymb}
\usepackage{subfigure}
\usepackage{subcaption}
\usepackage{array,booktabs,ragged2e}
\usepackage{multirow}
\usepackage{multicol}
\usepackage{times}
\usepackage[most]{tcolorbox}
\usepackage{latexsym}
\usepackage{arydshln}
\usepackage{hyperref}
\usepackage{pifont}
\usepackage{float}
\usepackage{bm}
\usepackage{microtype}
\usepackage{inconsolata}
\usepackage{xcolor}
\usepackage{xspace}
\usepackage{svg}

\usepackage{placeins}
\usepackage{afterpage}
\usepackage{amsfonts}
\usepackage{amsmath}

\usepackage[T1]{fontenc}
\usepackage[utf8]{inputenc}

\title{\textit{Nek Minit}: Harnessing Pragmatic Metacognitive Prompting for Explainable Sarcasm Detection of Australian and Indian English}

\author{Ishmanbir Singh*\quad Dipankar Srirag*\quad Aditya Joshi\\
University of New South Wales, Sydney, Australia\\
\quad\texttt{ishman.singh@student.unsw.edu.au,  $\{$d.srirag, aditya.joshi$\}$@unsw.edu.au} \\
* Equal contribution}

\begin{document}
\maketitle
\begin{abstract}
Sarcasm is a challenge to sentiment analysis because of the incongruity between stated and implied sentiment. The challenge is exacerbated when the implication may be relevant to a specific country or geographical region. Pragmatic metacognitive prompting (\textsc{pmp}) is a cognition-inspired technique that has been used for pragmatic reasoning. In this paper, we harness \textsc{pmp} for explainable sarcasm detection for Australian and Indian English, alongside a benchmark dataset for standard American English. We manually add sarcasm explanations to an existing sarcasm-labeled dataset for Australian and Indian English called \textsc{besstie}, and compare the performance for explainable sarcasm detection for them with \textsc{flute}, a standard American English dataset containing sarcasm explanations. Our approach utilising \textsc{pmp} when evaluated on two open-weight LLMs (\textsc{gemma} and \textsc{llama}) achieves statistically significant performance improvement across all tasks and datasets when compared with four alternative prompting strategies. We also find that alternative techniques such as agentic prompting mitigate context-related failures by enabling external knowledge retrieval.  The focused contribution of our work is utilising \textsc{pmp} in generating sarcasm explanations for varieties of English.

\end{abstract}

\section{Introduction}
Sarcasm is a form of verbal irony used to express contempt or ridicule, often by saying the opposite of what one means~\cite{joshi-sarcasm-survey}. This paper focuses on \textbf{explainable sarcasm detection as a generation task where, given a text, the LLM must predict if it contains sarcasm, and, if it does, it must generate a textual explanation}~\cite{kumar-etal-2022-become}. 
However, sarcasm is socio-culturally situated, and its interpretation often depends on local conventions~\cite{sarcasm-sociocultural-variables}.
This may be evidenced via phrases (such as `\textit{eshay}' or `\textit{nek minit}\footnote{\url{https://en.wikipedia.org/wiki/Nek\_minnit}; Accessed on 20th May 2025.} in the case of Australian English) or statements (such as `\textit{The sun is out and I am at work yay}' may not be understood as sarcastic by Indian English speakers). Recent work shows that large language models (LLMs) may misinterpret sarcasm in non-Western cultural contexts~\cite{atari_xue_park_blasi_henrich_2023, llm-cultural-bias}. Therefore, we focus on explainable sarcasm detection for two varieties of English: Australian (native variety) and Indian English (non-native variety), alongside standard American English. The extended discussion on the related works is provided in Appendix~\ref{sec:related-works}. 

We harness pragmatic metacognitive prompting (\textsc{pmp};~\citealp{lee-etal-2025-pragmatic}), an extension of metacognitive prompting~\cite{meta-cognitive-prompting}, to incorporate contextual understanding with respect to the geographical region for the task of explainable sarcasm detection. We compare \textsc{pmp} alongside four prompting baselines for a standard American English dataset (\textsc{flute};~\citealp{chakrabarty-etal-2022-flute}) and corresponding Australian and Indian English subsets from \textsc{besstie}~\cite{srirag2025besstiebenchmarksentimentsarcasm} using \textit{two} open-weight LLMs. Our proposed approach based on \textsc{pmp} significantly (\textit{p}$\leq$0.001) improves the performance on both tasks across all datasets\footnote{We provide the prompts used for our experiments in Appendix~\ref{sec:prompts}.}.  The novelty of this paper is three-fold: (a) We are the first to explore sarcasm explanation generation in the context of varieties of English; (b) we are the first to explore \textsc{pmp} and \textsc{kg} in the context of sarcasm explanation generation.; (c) we release a manually created dataset of sarcasm explanations for Australian and Indian English. Our findings corroborate past work that shows degradation in performance for language varieties other than standard American English for other tasks~\cite{blodgett-etal-2016-demographic, ziems2023multi, joshi2024natural, srirag-etal-2025-evaluating}.

\begin{figure*}[t!]
    \begin{adjustbox}{width=0.9\linewidth,center}
    \includegraphics{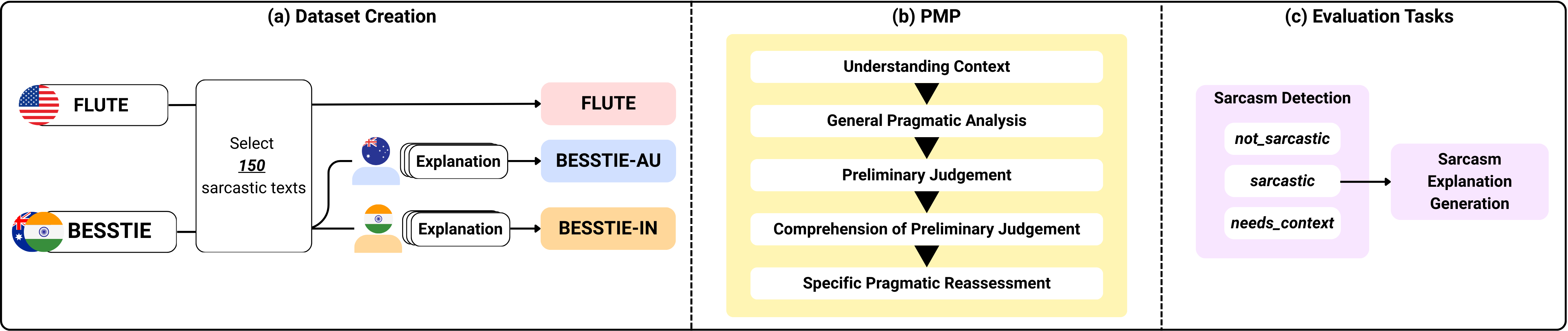}
    \end{adjustbox}
    \caption{Evaluation methodology; (a) Dataset Creation in Section~\ref{sec:extend}; (b) Prompting strategies in Section~\ref{sec:pmp-sec}; (c) Evaluation tasks in Section~\ref{sec:method}.}
    \label{fig:methodology-diagram}
\end{figure*}
\section{Proposed Approach}\label{sec:method}
Given an input text, our objective is explainable sarcasm detection. This spans two tasks: (a) sarcasm detection, and (b) the generation of a textual explanation when sarcasm is detected. Additionally, the model may output two auxiliary labels: \textit{not\_sarcastic}, if no sarcasm is found; and \textit{needs\_context}, if the model is uncertain and requires additional information to make a judgment. As the texts sampled are all labeled to have sarcasm, the expected output is an explanation for sarcasm for all test samples. Our evaluation methodology for the task is outlined in Figure~\ref{fig:methodology-diagram}.

\begin{table}[]
    \begin{adjustbox}{width=\linewidth,center}
    \scriptsize{
        \begin{tabular}{cccc}
        \toprule
            Subset & Samples & $\bar{\text{Text}}$ & $\bar{\text{Expl.}}$\\\midrule[\heavyrulewidth]
            \textsc{flute} & 150 & 19.1 & 28.2\\
            \textsc{besstie-au} & 150 & 47.7 & 25.7\\
            \textsc{besstie-in} & 150 & 14.3 & 29.6\\
        \bottomrule
        \end{tabular}
        }
    \end{adjustbox}
    \caption{Dataset statistics for the three evaluation subsets. The average length of the sarcastic text ($\bar{\text{Text}}$), and the average length of the corresponding ground truth explanation ($\bar{\text{Expl.}}$), are measured in words.}
    \label{tab:stats}
\end{table}

\subsection{Extending \textsc{\textsc{besstie}}}\label{sec:extend}
As shown in Figure~\ref{fig:methodology-diagram}(a), \textbf{we annotate 150 sarcastic examples each from Australian and Indian English subsets of \textsc{besstie} with text-based explanations.} Two authors of this paper, one native to Australia and one to India, served as annotators to independently write these sarcasm explanations for texts from their corresponding regional subset. Annotators were instructed to provide semi-structured explanations in a format stylistically aligned with \textsc{flute}. This enables consistency in explanation structure across all samples. We refer to these subsets as \textsc{besstie-au} and \textsc{besstie-in} respectively. We also randomly sample \textbf{150 sarcastic instances and their corresponding explanations} from \textsc{\textsc{flute}} as representative of standard American English. Table~\ref{tab:stats} reports basic statistics of the evaluation datasets, including the average length (in whitespace-tokenised words) of the sarcastic text and its explanation across the three subsets.

\subsection{\textsc{pmp} for Explainable Sarcasm Detection}\label{sec:pmp-sec}
As shown in Figure~\ref{fig:methodology-diagram}(b), we structure the prompt as a \textit{five-step} reasoning scaffold based on pragmatic linguistic theory~\cite{lee-etal-2025-pragmatic}\footnote{We provide an example output in the Appendix~\ref{sec:pmp}.}. While~\citet{lee-etal-2025-pragmatic} utilise \textsc{pmp} for sarcasm detection, we adapt \textsc{pmp} to sarcasm explanation generation for language varieties (Australian English and Indian English) as follows.
 \begin{enumerate}[noitemsep,nolistsep]
 \item \textbf{Comprehension of Context/Understanding}: The LLM identifies key situational elements that may trigger sarcastic meanings, enabling the deduction of intended meanings even for unfamiliar words or expressions.
 \item \textbf{General Pragmatic Analysis}: In this step, the LLM clarifies the true intent in the text by separating genuine beliefs from exaggerated or pretended attitudes. This can further clarify the meaning of unusual or unclear words or expressions.
\item \textbf{Preliminary Judgment}: An initial hypothesis explanation for sarcasm is generated as a focused seed to guide interpretation.
\item \textbf{Meta-Comprehension}: The LLM validates the hypothesis explanation by ensuring all relevant contextual cues (both explicit and implicit) are correctly interpreted.
\item \textbf{Specific Pragmatic Reassessment}: Finally, the LLM systematically examines detailed pragmatic elements, improving and justifying the final explanation, further resolving any residual ambiguity.
 \end{enumerate}

\begin{table*}[t!]
    \begin{adjustbox}{width=\linewidth,center}
        \begin{tabular}{cccccccccc}
        \toprule
            \multirow{2}{4em}{\centering Prompt} & \multicolumn{3}{c}{\textsc{flute}} & \multicolumn{3}{c}{\textsc{besstie-au}} & \multicolumn{3}{c}{\textsc{besstie-in}}\\\cmidrule(lr){2-4}\cmidrule(lr){5-7}\cmidrule(lr){8-10}
            & \textit{accuracy} & \textit{similarity} & \textit{judge} & \textit{accuracy} & \textit{similarity} & \textit{judge} & \textit{accuracy} & \textit{similarity} & \textit{judge}\\\midrule[\heavyrulewidth]
            \multicolumn{10}{c}{(a) \textsc{gemma}} \\\midrule[\heavyrulewidth]
            \textsc{zero} & 0.97 & 0.51 & 4.7 & 0.70 & 0.4 & 2.92 & 0.59 & 0.49 & 2.62 \\
            \textsc{few} & 0.98 & \textbf{0.63} & 4.74 & 0.73 & 0.42 & 2.99 & 0.59 & 0.49 & 2.55 \\
            \textsc{origin} & \texttt{-} & \texttt{-} & \texttt{-} & 0.72 & 0.42 & 3.12 & 0.61 & 0.52 & 2.78 \\
            \textsc{kg} & 0.93 & 0.45 & 4.47 & 0.74 & 0.4 & 2.99 & 0.85 & 0.59 & 3.31 \\ \hdashline
            \textsc{pmp} (Ours) & \textbf{1.0} & 0.5 & \textbf{4.92} & \textbf{0.94} & \textbf{0.49} & \textbf{3.98} & \textbf{0.91} & \textbf{0.63} & \textbf{3.76} \\\midrule[\heavyrulewidth]
            \multicolumn{10}{c}{(b) \textsc{llama}} \\\midrule[\heavyrulewidth]
            \textsc{zero} & 0.61 & 0.35 & 2.94 & 0.55 & 0.33 & 2.28 & 0.49 & 0.41 & 1.62 \\
            \textsc{few} & 0.69 & \textbf{0.43} & 3.22 & 0.35 & 0.26 & 1.66 & 0.42 & 0.39 & 1.43 \\
            \textsc{origin} & \texttt{-} & \texttt{-} & \texttt{-} & 0.57 & 0.34 & 2.15 & 0.55 & 0.41 & 1.69 \\
            \textsc{kg} & \textsc{n/a} & \textsc{n/a} & \textsc{n/a} & \textsc{n/a} & \textsc{n/a} & \textsc{n/a} & \textsc{n/a} & \textsc{n/a} & \textsc{n/a} \\\hdashline
            \textsc{pmp} (Ours) & \textbf{0.9} & 0.38 & \textbf{3.97} & \textbf{0.92} & \textbf{0.41} & \textbf{3.2} & \textbf{0.79} & \textbf{0.52} & \textbf{2.47} \\
        \bottomrule
        \end{tabular}
    \end{adjustbox}
    \caption{Performance comparison between baseline prompting methods and \textsc{pmp}, for tasks: sarcasm detection and sarcasm explanation generation. We test (a) \textsc{gemma} and (b) \textsc{llama} on the three datasets. We use \textit{accuracy} to measure the performance on sarcasm detection task, while \textit{similarity} and \textit{judge} are used for sarcasm explanation generation. \textsc{n/a} in (b) indicates \textsc{kg} was not applicable because it is not compatible with \textsc{llama}. The best task performances are denoted by numbers in \textbf{bold}.}
    \label{tab:explanation}
\end{table*}

\section{Experiment Setup}
We experiment with \textit{two} open-weight decoder-only model optimised for instruction following, namely, Gemma3-12B-Instruct (\textsc{gemma};~\citealp{gemmateam2025gemma3technicalreport}) and Llama-3.2-3B-Instruct (\textsc{llama};~\citealp{grattafiori2024llama3herdmodels})\footnote{Text Generation Settings: max\_new\_tokens=1024, temperature=1.0, top\_p=0.95, top\_k=64}. All experiments are performed using \textit{one} A100 GPU. We compare \textsc{pmp} with \textit{four} baseline methods:
\begin{enumerate}[noitemsep,nolistsep]
\item \textbf{\textsc{zero}}: We prompt the model directly with a task instruction for sarcasm detection and explanation, without any examples.
\item \textbf{\textsc{few}}: We include five manually curated examples from \textsc{flute} to the \textsc{zero} prompt. Each example contains a sarcastic sentence paired with an explanation, and these examples are fixed across all evaluations.
\item \textbf{\textsc{origin}}: We append the geographical origin to the \textsc{zero} prompt to help the model incorporate cultural and linguistic cues specific to the source variety.
\item \textbf{\textsc{kg}}: We implement an agentic prompting strategy inspired by ReAct~\cite{yao2023react}. The model is prompted to identify knowledge gaps, integrate information retrieved through search queries, and then respond. The model uses the DuckDuckGo tool\footnote{\url{https://github.com/deedy5/duckduckgo_search}; Accessed on 19 May 2025} to issue search queries. This method of prompting minimises the gap in context and knowledge from local culture, idioms, or events\footnote{An example output is provided in the Appendix~\ref{sec:kg}.}.
\end{enumerate}
We measure performance across three metrics: (a) \textit{accuracy} (the proportion of correctly predicted labels over all instances); (b) \textit{similarity} (average cosine similarity between the Sentence-BERT embeddings of the reference explanation and generated explanation;~\citealp{reimers2019sentencebert}); and \textit{judge}. For \textit{judge}, we employ GPT-4o~\cite{openai2024gpt4technicalreport} with default parameters as an evaluator to assess explanation quality. Given a sarcastic text sample, the corresponding ground truth explanation, and the generated explanation, GPT-4o assigns a score from 0 to 5 based on an arbitrary scoring criterion defined in Appendix~\ref{sec:score}. While \textit{accuracy} measures the sarcasm detection performance, \textit{similarity} and \textit{judge} evaluate the generated explanations. 
\section{Results}
Our results address three questions: (a) How do baseline prompting strategies perform on sarcasm detection and explanation generation?; (b) Does \textsc{pmp} improve over these strategies, particularly for sarcasm explanation task?; and (c) Under what settings does \textsc{pmp} yield the best performance?

Table~\ref{tab:explanation} reports results across the \textit{three} datasets for the \textit{two} tasks. For both models, \textsc{zero} and \textsc{few} yield high results on \textsc{flute} for sarcasm detection, with \textsc{gemma} achieving 0.97 and 0.98 \textit{accuracy} respectively, and \textsc{llama} yielding 0.61 and 0.69. However, these methods degrade substantially on the \textsc{besstie} subsets. Particularly, the performance of \textsc{gemma} when tested on \textsc{besstie-in} remains at 0.59 for both methods, while \textsc{llama} shows a slight degradation (\textsc{zero}: 0.49; \textsc{few}: 0.42). Appending geographic cues in \textsc{origin} prompt yields inconsistent improvements when compared to \textsc{zero}. For example, \textsc{gemma} reports a minor improvement when tested on \textsc{besstie-au} (\textsc{zero}: 0.7; \textsc{origin}: 0.72), \textsc{llama} reports a higher performance gain when tested on \textsc{besstie-in} (\textsc{zero}: 0.49; \textsc{origin}: 0.55). These findings indicate that appending geographical cues alone is insufficient to capture cultural nuances. For explanation generation, the trends among baselines remain consistent. \textsc{gemma}, when prompted with \textsc{kg}, demonstrates improvements in explanation quality on \textsc{besstie-in} (\textit{similarity}: 0.59; \textit{judge}: 3.31), suggesting some benefit from agentic reasoning. However, these gains do not consistently transfer across datasets. Moreover, \textsc{kg} is not compatible with \textsc{llama}, limiting its applicability. 

\begin{table}[]
  \begin{adjustbox}{width=\linewidth,center}
    \begin{tabular}{ccccccc}
    \toprule
      \multirow{2}{4em}{\centering Prompt} & \multicolumn{2}{c}{\textsc{flute}} & \multicolumn{2}{c}{\textsc{besstie-au}} & \multicolumn{2}{c}{\textsc{besstie-in}}\\\cmidrule(lr){2-3}\cmidrule(lr){4-5}\cmidrule(lr){6-7}
        & \textsc{ns} & \textsc{nc} & \textsc{ns} & \textsc{nc} & \textsc{ns} & \textsc{nc} \\\midrule[\heavyrulewidth]
      \textsc{zero} & 4 & 0 & 36 & 9 & 18 & 43 \\
      \textsc{few} & 2 & 0 & 31 & 10 & 15 & 46 \\
      \textsc{origin} & - & - & 30 & 10 & 21 & 36 \\
      \textsc{kg} & 8 & 0 & 36 & 1 & 18 & 4 \\ \hdashline
      \textsc{pmp} (Ours) & 0 & 0 & 4 & 2 & 1 & 13 \\
    \bottomrule
    \end{tabular}
  \end{adjustbox}
  \caption{Counts of sarcastic instances, across datasets and prompting strategies, flagged by \textsc{gemma} to require more context (\textsc{nc}) or incorrectly identified to not present sarcasm (\textsc{ns}).}
  \label{tab:error}
\end{table}

\textsc{pmp} significantly improves over all baselines across both tasks and models. For \textsc{gemma}, \textsc{pmp} achieves the highest detection accuracy (1.00) and explanation quality (\textit{judge}: 4.92) on \textsc{flute}, and similarly strong performance on \textsc{besstie-au} (\textit{accuracy}: 0.94; \textit{judge}: 3.98) and \textsc{besstie-in} (\textit{accuracy}: 0.91; \textit{judge}: 3.76). For \textsc{llama}, \textsc{pmp} significantly improves over baseline performances: on \textsc{besstie-in}, it improves \textit{similarity} and \textit{judge} to 0.52 and 2.47 compared to \textsc{origin}(0.41 and 1.69), and \textit{accuracy} of 0.79 from 0.55. We also note that while each row of the table reports lower values for Australian and Indian English as compared to standard American English present in \textsc{flute}, our results on \textsc{pmp} method highlight the potential to bridge the performance gap between varieties. The improvements across all datasets are statistically significant (p$\leq$0.001) relative to the \textsc{zero} baseline.

Table~\ref{tab:error} presents an error analysis of two types of explanation errors: instances incorrectly classified as \textsc{ns}, and those requiring \textsc{nc}. On both \textsc{besstie} datasets, \textsc{pmp} substantially reduces these errors. On \textsc{besstie-in}, the count of \textsc{ns} cases drops from 18 under \textsc{zero} to 1 with \textsc{pmp}; similarly, \textsc{nc} cases reduce from 43 to 13. These results are consistent for \textsc{besstie-au}, confirming that \textsc{pmp} is particularly effective in cases requiring nuanced cultural or pragmatic interpretation. However, for \textsc{besstie-in}, models more often output \textsc{nc}, especially for \textsc{zero} and \textsc{few}. This indicates that the model often cannot interpret the sarcasm from surface-level language alone, likely due to cultural or regional differences in how sarcasm is expressed. Here, \textsc{kg} proves effective by supplying external context that helps the model bridge this gap, reducing the \textsc{ns} errors to 4, when compared to 43 from \textsc{zero}. 
\section{Conclusion}
We demonstrated how pragmatic metacognitive prompting (\textsc{pmp}) can be harnessed to generate sarcasm explanations for text written in varieties. We evaluate our approach on three language varieties of English: standard American English, Australian and Indian English, and two open-source LLMs: \textsc{gemma} and \textsc{llama}. We annotated an existing dataset for the latter two with sarcasm explanations to assess model performance for explainable sarcasm detection. Standard prompting methods, such as zero-shot and few-shot, perform well on \textsc{flute}, but failed to generalise to Australian and Indian English subsets of \textsc{besstie}. \textbf{\textsc{pmp} significantly (p$\leq$0.001) improved the performance across both tasks, both Australian and Indian English, and both the models}. Agentic prompting methods like \textsc{kg} also reduced context-related failures by enabling dynamic knowledge integration. Our results and error analysis demonstrated the limitations of generic prompts and the importance of pragmatic scaffolding for figurative language understanding. Our findings suggest that reasoning-aware prompting offers a viable pathway to improve explanation generation for sarcasm.

\section*{Limitations}
We perform sarcasm detection only on positive text samples, i.e. sarcastic text, as this study primarily explores sarcasm explanation as a task. There may be multiple possible explanations for sarcasm. Also, we acknowledge that there are varieties within a country related to the state, native language, and so on. Note that we perform sarcasm detection only on positive text samples, i.e. sarcastic text, as this study primarily explores sarcasm explanation as a task. However, as a first study on sarcasm explanation for language varieties, our observations are a good starting point for the future. The proposed \textsc{pmp} approach helps \textsc{besstie-au} and \textsc{besstie-in} to retrieve appropriate context which is not necessary in the case of \textsc{flute} based on past findings in the bias of LLMs towards a western-centric context.

\section*{Ethical Considerations}
The research was approved by the ethics board of the host institution. All annotators were native speakers of the respective English varieties (Australian and Indian) and participated voluntarily in the annotation process. Given that sarcasm is often context-dependent and culturally embedded, we acknowledge the sensitivity involved in interpreting or misclassifying user-generated content. No personally identifiable information was used in this study.
\section*{Acknowledgment}
This paper is the outcome of a Taste-of-Research scholarship awarded to Ishmanbir Singh by the Faculty of Engineering at UNSW Sydney. The paper is dedicated to the memory of Pushpak Bhattacharyya from Indian Institute of Technology Bombay who served as the ACL President (2016), and was considered a pioneer of NLP in India.
\bibliography{refs}
\onecolumn
\appendix
\section{Related Work}\label{sec:related-works}
Sarcasm detection has employed statistical models~\cite{joshi-sarcasm-survey}, sentiment-incongruity heuristics~\cite{riloff-etal-2013-sarcasm}, and traditional neural architectures such as LSTMs~\cite{ghosh-etal-2017-role}, CNNs~\cite{poria-etal-2016-deeper}, GNNs~\cite{10.1145/3404835.3463061,liang-etal-2022-multi} and Transformer~\cite{yao2024sarcasmdetectionstepbystepreasoning}. \textbf{Prompting LLMs} has shown promise in guiding models to reason about sarcasm effectively ~\cite{liu-etal-2023-prompt, Yao_Zhang_Li_Qin_2025}. Specifically, Pragmatic Metacognitive Prompting (\textsc{pmp})~\cite{lee-etal-2025-pragmatic} introduces a structured approach that mirrors human pragmatic reasoning by incorporating reflection and analysis of implied meanings, contextual cues, and speaker intent. This method has demonstrated improved performance in sarcasm detection tasks but \textbf{has not been used for sarcasm explanation generation}. Additionally, agentic prompting methods like ReAct~\cite{yao2023react} enable models to actively retrieve and integrate external knowledge, facilitating context-aware reasoning in sarcasm detection. Beyond sarcasm detection, \textbf{sarcasm explanation} generation has been investigated using datasets like \textsc{flute}~\cite{chakrabarty-etal-2022-flute} which provide figurative language instances with explanations. Models such as TEAM~\cite{jing-etal-2023-multi} employ multi-source semantic graphs to generate multimodal sarcasm explanations, integrating visual and textual cues. Ours is the first work that examines the different prompting strategies for the explanation generation of sarcasm for language varieties.

\section{Prompts}\label{sec:prompts}
In this paper, we evaluate models on the task of explainable sarcasm detection, by prompting models to first identify sarcasm and then provide an explanation for classifying the text to be sarcastic. Below are the prompts that we used.
\subsection{\textsc{zero}}
    \texttt{For the provided text, perform one of the tasks. If the text is not sarcastic, output `not\_sarcastic'. If the text is sarcastic, provide an explanation in one or two sentences. Output `need\_context' if you cannot explain the sarcasm.}
\subsection{\textsc{few}}

    \texttt{For the provided text, perform one of the tasks. If the text is not sarcastic, output `not\_sarcastic'. If the text is sarcastic, provide an explanation in one or two sentences. Output `need\_context' if you cannot explain the sarcasm. Following are some examples of sarcastic texts and their explanations.}
\texttt{\begin{enumerate}
    \item \textbf{Text}: I got the greatest news today when I found out my company is getting rid of me after 10 years of faithful service\\
Explanation: sarcastic. Being fired from a job after 10 years of service is not something to be happy about and hence receiving the greatest news is sarcastic.
    \item \textbf{Text}: I feel great that I am single and have not accomplished anything significant inspite of being 30 years old\\
\textbf{Explanation}: sarcastic. Someone who is 30 years old and single has not achieved anything significant in their life so far and so should not feel honored by that fact.
    \item \textbf{Text}: The idea of the genius Republican party blatantly rejecting climate change and its effects on our planet makes me feel safe for the future\\
\textbf{Explanation}: sarcastic. The Republican party's denial of climate change will have catastrophic effects on our planet in the future and so to feel confident about the future is absurd.
    \item \textbf{Text}: I was so impressed when my hair dresser tried a new product on my hair but the end result was me getting to shave all my hair off\\
\textbf{Explanation}: sarcastic. Shaving all your hair off is not an impressive hairstyle and is often seen as a last resort when someone has messed up your hair.
    \item \textbf{Text}: I am on seventh heaven because I don't have enough money to pay my important bill.\\
\textbf{Explanation}: sarcastic. Having not enough money to pay an important bill creates a lot of stress and anxiety and is definitely not a heavenly feeling.
\end{enumerate}}

\subsection{\textsc{origin}}
\texttt{This text is from [Australian/Indian] subreddit post or comment. If the text is not sarcastic, output `not\_sarcastic'. If the text is sarcastic, provide an explanation in one or two sentences. Output `need\_context' if you cannot explain the sarcasm.}
\subsection{\textsc{kg}}
\texttt{Think step-by-step and feel free to use tools whenever they would be helpful.
Once you have finished reasoning and using tools:}
\texttt{
\begin{itemize}
    \item If the text is not sarcastic:\\
  \{``action'': ``Final Answer'', ``action\_input'': ``not\_sarcastic''\}
    \item If the text is sarcastic:\\
  \{``action'': ``Final Answer'', ``action\_input'': ``sarcastic. Explanation: concise explanation (1-2 sentences)''\}
  \item If you cannot explain the sarcasm:\\
  \{``action'': ``Final Answer'', ``action\_input'': ``need\_context''\}
\end{itemize}
All outputs must be valid JSON, with no unescaped double quotes; use single quotes. You have access to the following tools:
\begin{itemize}
    \item \textbf{Search}: Tool for getting up to date answers to current or historical events and word/phrase definitions., args: \{`tool\_input': \{`type': `string'\}\}
\end{itemize}
Use a json blob to specify a tool by providing an action key (tool name) and an action\_input key (tool input). Valid ``action'' values are: ``Final Answer'' or ``Search''.\\
\\
Provide only ONE action per \$JSON\_BLOB, as shown:\\
\{``action'': \$TOOL\_NAME, ``action\_input'': \$INPUT\}\\
\\
Follow this format:\\
\\
Question: input question to answer\\
Thought: consider previous and subsequent steps\\
Action:
\$JSON\_BLOB\\
Observation: action result\\
... (repeat Thought/Action/Observation N times)\\
Thought: I know what to respond\\
Action:\\
\{``action'': ``Final Answer'', ``action\_input'': ``Final response to human''\}\\
\\
Begin! Reminder to ALWAYS respond with a valid json blob of a single action. Use tools if necessary. Respond directly if appropriate. Format is\\
Action:\$JSON\_BLOB then Observation:.\\
Thought:\\
Human:
}
\begin{tcolorbox}[
  colback=white, colframe=black, title=Additional Information,
  fonttitle=\bfseries, coltitle=white]
Once the model outputs a thought and action, the action is executed and the thought, action and output are all appended back into the prompt. This repeats until the "Final Answer" action is output by the model.
\end{tcolorbox}
\subsection{\textsc{pmp}}
\texttt{For the provided text, perform one of the tasks. If the text is not sarcastic, output `not\_sarcastic'. If the text is sarcastic, follow these steps in your analysis:}
\texttt{\begin{enumerate}
    \item \textbf{Comprehension of Context/Understanding:}
    \begin{itemize}
       \item Repeat the text and summarize it to confirm your understanding
       \item Identify the context, topic, and key elements of the situation described
     \end{itemize}
    \item \textbf{General Pragmatic Analysis:}
    \begin{itemize}
       \item What does the author imply about the situation with their statement?
       \item What does the author think about the situation?
       \item Are what the author implies and what they think saying the same thing?
       \item Is the author pretending to have a certain attitude toward the conversation?
     \end{itemize}
    \item \textbf{Preliminary Judgment:}
    \begin{itemize}
       \item Based on your analysis, formulate an initial explanation of why the text is sarcastic
     \end{itemize}
    \item \textbf{Comprehension of Preliminary Judgment/Context:}
    \begin{itemize}
       \item Summarize your preliminary analysis
       \item Ensure you've captured all relevant contextual elements
       \end{itemize}
    \item \textbf{Specific Pragmatic Analysis/Reassessment:}
    \begin{itemize}
       \item Implicature: What is implied beyond the literal meaning?
       \item Presuppositions: What information is taken for granted?
       \item Intent: What does the author hope to achieve with their statement?
       \item Polarity: Is there a mismatch between apparent and intended tone?
       \item Pretense: Is there pretense in the author's attitude?
       \item Meaning gap: What is the difference between literal and implied meaning?
       \item Context: Consider slang, cultural references, or communication patterns
       \end{itemize}
    \item \textbf{After completing your analysis, perform one of the tasks.} 
    \begin{itemize}
       \item If the text is sarcastic, provide an explanation in one or two sentences.
       \item Output `need\_context' if you cannot explain the sarcasm.
       \end{itemize}
\end{enumerate}}

\section{\textsc{pmp} Example Output}\label{sec:pmp}

\begin{tcolorbox}[
  colback=white, colframe=black, title=\textsc{pmp} Output Analysis,
  fonttitle=\bfseries, coltitle=white]

\small

\begin{tcolorbox}[
  colback=teal!10!white, colframe=teal!50!black, title=Input Text, boxsep=1pt,left=2mm, right=2mm]
\textbf{Text:} Whitlam wanted to shut it down did he not? Nek Minit
\end{tcolorbox}

\begin{tcolorbox}[
  colback=gray!5!white, colframe=gray!75!black,
  title=Step 1: Comprehension of Context/Understanding, boxsep=1pt,left=2mm, right=2mm]
\textbf{Summary:} The text poses a rhetorical question about Whitlam's desire to shut something down, followed by the phrase \textit{Nek Minit}, which is Australian slang meaning \textit{in a short time} or \textit{very soon}.\\
\textbf{Context:} he context appears to be Australian politics, referencing Gough Whitlam, a former Prime Minister of Australia. The ``it'' refers to something he allegedly wanted to shut down. The inclusion of \textit{Nek Minit} suggests a playful or dismissive response to that allegation.
\end{tcolorbox}

\begin{tcolorbox}[
  colback=gray!5!white, colframe=gray!75!black,
  title=Step 2: General Pragmatic Analysis, boxsep=1pt,left=2mm, right=2mm]
\textbf{Author's Implied Meaning:} The author likely implies that the idea of Whitlam wanting to shut something down is overstated or a common, possibly exaggerated, claim.\\
\textbf{Author's Attitude:} The author's attitude is likely skeptical or humorous. They are not taking the claim seriously.\\
\textbf{Alignment of Implied and Stated:} The implied meaning and the stated question don't perfectly align. The question \textit{seems} to solicit confirmation, but the \textit{Nek Minit} undermines that, suggesting the matter is trivial or quickly resolved.\\
\textbf{Pretense of Attitude:} The author is pretending to be earnestly seeking confirmation, but the \textit{Nek Minit} reveals a dismissive or playful attitude.
\\
\textbf{Pretense:} Appears serious but sets up mockery.
\end{tcolorbox}

\begin{tcolorbox}[
  colback=gray!5!white, colframe=gray!75!black,
  title=Step 3: Preliminary Judgment, boxsep=1pt,left=2mm, right=2mm]
The text is likely sarcastic because the rhetorical question is followed by a phrase (\textit{Nek Minit}) that minimizes the significance of the issue being questioned, creating a mismatch between the apparent seriousness of the question and the implied triviality.
\end{tcolorbox}

\begin{tcolorbox}[
  colback=gray!5!white, colframe=gray!75!black,
  title=Step 4: Comprehension of Preliminary Judgment/Context, boxsep=1pt,left=2mm, right=2mm]
The text employs a rhetorical question about Whitlam's actions within the context of Australian politics. The combination of the question and \textit{Nek Minit} creates a sarcastic effect by downplaying the significance of Whitlams alleged desire to shut something down, suggesting the issue is either minor or quickly resolved.
\end{tcolorbox}

\begin{tcolorbox}[
  colback=gray!5!white, colframe=gray!75!black,
  title=Step 5: Specific Pragmatic Analysis/Reassessment, boxsep=2pt,left=2mm, right=2mm]
\textbf{Implicature:} The implicature is that the claim about Whitlam's actions is not a major concern or deserves lengthy consideration.\\
\textbf{Presuppositions:} It presupposes that the audience is familiar with Gough Whitlam and potentially has some knowledge of the issue being referenced \\(whatever ""it"" is).  It also presupposes an understanding of \textit{Nek Minit} as slang indicating immediacy or triviality.\\
\textbf{Intent:} The author intends to express skepticism or amusement about the claim in a playful, culturally specific way.\\
\textbf{Polarity:} Theres a significant polarity mismatch. The question has a formal, interrogative structure (suggesting seriousness), while \textit{Nek Minit} introduces an informal, dismissive tone.\\
\textbf{Pretense:} The author is feigning serious inquiry while actually expressing a dismissive attitude.\\
\textbf{Meaning Gap:} The literal meaning of the question is a request for confirmation. The implied meaning is a casual rejection of the claims importance.\\
\textbf{Context:} The use of Australian slang (\textit{Nek Minit}) is crucial to understanding the sarcasm.  It's a culturally specific expression that signals informality and downplaying.
\end{tcolorbox}

\begin{tcolorbox}[
  colback=lime!10!white, colframe=lime!50!black,
  title=Step 6: Final Explanation, boxsep=1pt,left=2mm, right=2mm]
The rhetorical question regarding Whitlams alleged action is undermined by the casual Australian slang \textit{Nek Minit}, creating a sarcastic effect by suggesting the issue is trivial and doesnt warrant serious consideration.
\end{tcolorbox}

\begin{tcolorbox}[
  colback=green!10!white, colframe=green!50!black,
  title=Ground Truth Explanation, boxsep=1pt,left=2mm, right=2mm]
The comment seems supportive of Prime Minister Whitlam but the Nek Minit (suggesting an ironic twist of events) pokes fun at how he was dismissed from government before being able to do anything.
\end{tcolorbox}

\end{tcolorbox}
\section{\textsc{kg} Example Output}\label{sec:kg}

\begin{tcolorbox}[colback=white, colframe=black, title=\textsc{kg} Reasoning Chain, fonttitle=\bfseries, coltitle=white, before skip=5pt, after skip=5pt
  ]
  \small
  \begin{tcolorbox}[colback=teal!10!white, colframe=teal!50!black, title=Input Text,left=2mm, right=2mm]
  \textbf{Text:} This case seriously is now sounding like a badly written \textit{Crime Patrol} episode.
  \end{tcolorbox}


  \begin{tcolorbox}[colback=gray!5!white, colframe=gray!75!black, title=Thought 1,left=2mm, right=2mm]
  The text uses a comparison to a low-quality, fictional crime show (\textit{Crime Patrol}) to express disbelief or mockery of a real-life case. The phrase ``badly written'' implies that the case is illogical, predictable, or absurd, but instead of stating this directly, it uses the humorous comparison to a poorly written TV show. This indirect, mocking comparison is characteristic of sarcasm. I need to check what \textit{Crime Patrol} is to ensure the comparison is understood.
  \end{tcolorbox}


  \begin{tcolorbox}[colback=yellow!10!white, colframe=yellow!50!black, title=Action,left=2mm, right=2mm]
  \textbf{Action:} Search \\
  \textbf{Action Input:} \texttt{"What is Crime Patrol?"}
  \end{tcolorbox}


  \begin{tcolorbox}[colback=orange!10!white, colframe=orange!50!black, title=Action Output,left=2mm, right=2mm]
  \textit{Crime Patrol}, the iconic true-crime anthology series hosted by Anup Soni, is making its way to Netflix on March 17, 2025. The series, which premiered in 2003, is the longest-running reality crime television series in India. \textit{Crime Patrol} not only tells such gripping stories but also teaches important lessons about staying safe and preventing crime. By showing the efforts of law enforcement, the show helps audiences see the importance of being aware and cautious in today's world.
  \end{tcolorbox}


  \begin{tcolorbox}[colback=gray!5!white, colframe=gray!75!black, title=Thought 2,left=2mm, right=2mm]
  The search result confirms that \textit{Crime Patrol} is a well-known Indian crime show, known for its dramatized recreations of real crimes. Given that the original statement compares the case to ``a badly written \textit{Crime Patrol} episode,'' it suggests the case being described is convoluted, unrealistic, or predictable—much like a poorly constructed fictional crime narrative. The sarcasm lies in the unexpected and humorous comparison to a show known for its sometimes sensationalized and dramatic portrayal of crime, implying the real-life case is far less sophisticated than it should be.
  \end{tcolorbox}


 \begin{tcolorbox}[colback=lime!10!white, colframe=lime!50!black, title=Final Output Explanation,left=2mm, right=2mm]
 The statement uses a negative comparison to \textit{Crime Patrol} to mock the quality of the case, implying it is unrealistic or poorly handled.
 \end{tcolorbox}
 \begin{tcolorbox}[colback=green!10!white, colframe=green!50!black, title=Ground Truth Explanation,left=2mm, right=2mm]
 The user sarcastically compares the case to a poorly written episode of \textit{Crime Patrol}, a popular Indian crime reenactment show—implying that the developments in the case are so bizarre, dramatic, or unbelievable that they feel fictional and poorly scripted, rather than real-life events.
 \end{tcolorbox}
\end{tcolorbox}

\newpage
\section{Scoring Criterion}\label{sec:score}
\small
\begin{tcolorbox}[colback=gray!5!white, colframe=black!40!white, title=Scoring Criterion, label=scoring-criteria]

\begin{tcolorbox}[colback=red!10!white, colframe=red!50!black, boxrule=0.5pt, left=2pt, right=2pt, top=2pt, bottom=2pt, enhanced]
\textbf{0:} {Irrelevant or incoherent explanation}
\end{tcolorbox}

\begin{tcolorbox}[colback=orange!10!white, colframe=orange!50!black, boxrule=0.5pt, left=2pt, right=2pt, top=2pt, bottom=2pt, enhanced]
\textbf{1:} {Barely related, vague or generic statement}
\end{tcolorbox}

\begin{tcolorbox}[colback=yellow!10!white, colframe=yellow!50!black, boxrule=0.5pt, left=2pt, right=2pt, top=2pt, bottom=2pt, enhanced]
\textbf{2:} {Somewhat related but incomplete or unclear reasoning}
\end{tcolorbox}

\begin{tcolorbox}[colback=lime!10!white, colframe=lime!50!black, boxrule=0.5pt, left=2pt, right=2pt, top=2pt, bottom=2pt, enhanced]
\textbf{3:} {Reasonable explanation, covers core sarcastic cue}
\end{tcolorbox}

\begin{tcolorbox}[colback=green!10!white, colframe=green!50!black, boxrule=0.5pt, left=2pt, right=2pt, top=2pt, bottom=2pt, enhanced]
\textbf{4:} {Strong explanation with appropriate contextual grounding}
\end{tcolorbox}

\begin{tcolorbox}[colback=teal!10!white, colframe=teal!50!black, boxrule=0.5pt, left=2pt, right=2pt, top=2pt, bottom=2pt, enhanced]
\textbf{5:} {Excellent explanation, highly aligned with human interpretation}
\end{tcolorbox}

\end{tcolorbox}
\end{document}